\documentclass[10pt,letterpaper]{article}

\usepackage{cogsci}

\cogscifinalcopy 
\usepackage{xcolor}
\usepackage{amsmath}
\usepackage{amssymb}
\usepackage{amsfonts}
\usepackage{graphicx}
\usepackage{pslatex}
\usepackage{apacite}
\usepackage{float} 

\newif\ifcomments
\commentstrue

\newcommand\I{\ensuremath{\mathbf{I}}}
\newcommand\C{\ensuremath{\mathbf{c}}}
\newcommand\U{\ensuremath{\mathbf{U}}}
\newcommand\fs{\ensuremath{f_{S}}}
\newcommand\fl{\ensuremath{f_{L}}}

\DeclareMathOperator*{\argmax}{arg\,max}

\title{Learning to refer informatively by amortizing pragmatic reasoning}

\author{{\large \bf Julia White\textsuperscript{1}, Jesse Mu\textsuperscript{2}, Noah D.~Goodman\textsuperscript{2,3}} \\ 
    Departments of \textsuperscript{1}Electrical Engineering, \textsuperscript{2}Computer Science, and \textsuperscript{3}Psychology \\
    Stanford University \\
    \texttt{\{jiwhite,muj,ngoodman\}@stanford.edu}}

\begin{document}

\maketitle

\begin{abstract}
A hallmark of human language is the ability to effectively and efficiently convey contextually relevant information. One theory for how humans reason about language is presented in the Rational Speech Acts (RSA) framework, which captures pragmatic phenomena via a process of recursive social reasoning \cite{Goodman2016}. However, RSA represents ideal reasoning in an unconstrained setting. We explore the idea that speakers might learn to \emph{amortize} the cost of RSA computation over time by directly optimizing for successful communication with an internal listener model. In simulations with grounded neural speakers and listeners across two communication game datasets representing synthetic and human-generated data, we find that our amortized model is able to quickly generate language that is effective and concise across a range of contexts, without the need for explicit pragmatic reasoning.

\end{abstract}

\section{Introduction}
When we refer to an object or situation using natural language, we easily generate an utterance that achieves a useful amount of information in the given context.
Counterfactual reasoning about alternative utterances has been central to explanations of these pragmatic aspects of language
\cite{Grice1975,searle1969speech}.
Yet these theories entail computations that appear slow and burdensome as cognitive processes, so how do we produce pragmatic utterances so easily?

One popular computational account of pragmatic reasoning is the Rational Speech Acts (RSA) model \cite{Goodman2016}.
In RSA, speakers and listeners reason about the meaning of utterances by considering the other utterances that could have been produced,
thus arriving at pragmatic interpretations that enrich literal semantics. In addition to referring expression generation \cite{frank2012predicting,degen2020when}, RSA has seen success in modeling a wide variety of phenomena, including scalar implicature, metaphor, hyperbole, and politeness (for a review, see \citeNP{Goodman2016}).


However, the successes of RSA have been in describing pragmatic language \textit{competence}. In other words, within Marr's \citeyear{marr1976understanding} levels of analysis,
RSA is a \emph{computational} account of human language,
describing only \emph{what} is to be ideally computed, and not \emph{how} humans produce language with limited resources.
Indeed, the full computation specified by RSA involves reasoning over all possible utterances in context, which is intractable in all but the most trivial settings. Moreover, such computation is wasteful in that it is \emph{memoryless}, and does not leverage past experience \cite{gershman2014amortized}.

The impracticality of explicit pragmatic reasoning has had implications for both theory and modeling.
Linguists have argued that pragmatic phenomena such as scalar implicature
\cite{levinson2000presumptive}
and metaphor \cite{lakoff2008metaphors} are \emph{conventionalized}, 
becoming the default interpretation of utterances unless specifically cancelled by the speaker.
Others have argued that during pragmatic reasoning, listeners may only consider a subset of relevant interpretations \cite{sperber1986relevance,fox2011characterization}, or trade off between ``slow'' and ``fast'' processes \cite{degen2019constraint}.
In natural language processing,
RSA-based models for pragmatic referring expression generation use approximate inference methods,
either sampling a subset of possible utterances from a learned model \cite{andreas2016reasoning,Monroe2017} or reasoning incrementally \cite{Cohn-Gordon2019}. Others use heuristics inspired from the psycholinguistics literature, with greedy and probabilistic search techniques to reduce the space of possible utterances \cite{krahmer2012computational,van2019conceptualization}.

\begin{figure}[tb]
\begin{center}
\includegraphics[width=8cm]{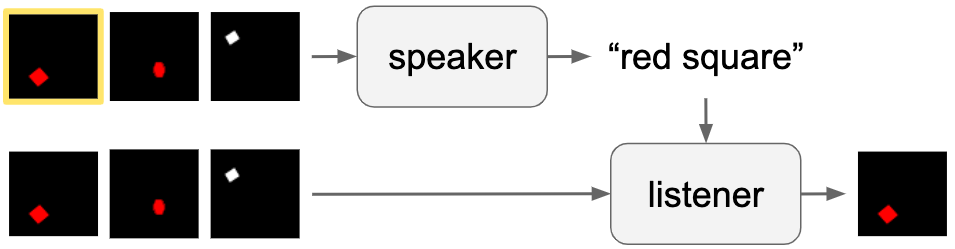}
\end{center}
\caption{Reference game: Three images including one target (indicated by a gold box) are given to a speaker and listener. The speaker produces an utterance for the listener who then must choose the target image. In RSA, the speaker explicitly reasons about the listener's interpretation of alternative utterances; in our amortized model, the speaker is trained to generate informative utterances without such reasoning.}
\label{example}
\end{figure}

Given limited time and resources, it is indeed unlikely that we do exhaustive Bayesian reasoning every time we speak.
An alternative is that experience with this slower reasoning process has led to amortized ``habits of speech'' that we use most of the time \cite{gershman2014amortized}.
In this paper, we model this hypothesis through grounded language models that are trained with listeners in Lewis \citeyear{Lewis1969}-style communication games (Fig.~\ref{example}),
measuring the degree to which they learn to speak pragmatically.
Our models are able to acquire desirable pragmatic behavior while being far more efficient than naive RSA-based approaches.
This suggests that agents can internalize pragmatic language production processes, providing a plausible \emph{algorithmic} account of human language production in resource-constrained settings
and a promising avenue for the development of more efficient pragmatic language models.

\section{How to Speak Informatively}

We explore communication within Lewis-style signalling games \citeyear{Lewis1969} which allow us to analyze language use in grounded contexts. Formally, a reference game $(\I, t)$ consists of a set of $N$ images $\I = (I_1, \dots, I_n)$, with one target image $I_t$ known only to the speaker. The speaker's job is to produce an utterance $u$ which, when given to the listener, causes the listener to correctly identify $t$ (Fig.~\ref{example}). Crucially, the ideal language for image $I_t$ changes depending on context. For example, in Fig.~\ref{example}, it is not sufficient to simply say the color (``red'') or shape (``square'') of the target image; both must be provided to be unambiguous.

We consider a variety of speakers in this setting, each capturing different cognitive hypotheses about utterance production. The baseline \textbf{naive speaker} is a neural captioning model trained to generate observed descriptions of an image, completely ignoring context. To incorporate pragmatics, the naive speaker can be provided context objects as input, which we call the \textbf{contextual speaker}. 
In contrast, \textbf{RSA-based speakers} explicitly consider alternative utterances: the gold-standard \textbf{Full RSA} model does the complete reasoning process specified by RSA theory; the \textbf{sample-rerank} model approximates RSA
by using a naive speaker to generate candidate utterances,
reducing the space of alternatives.
Our \textbf{amortized} speaker takes a different approach, training a context-aware language model with the RSA communication objective, without any pragmatic reasoning.
The model is trained directly via backpropagation through a (white-box) internal listener model.
In contrast, we also consider a \textbf{REINFORCE} speaker which learns from a sparser communicative reward signal given by a (black-box) external listener.

\paragraph{Naive speaker.} The \textbf{naive speaker} $S_\text{0}$ is a standard image captioning (IC) model trained to generate referring expressions that completely ignore context. Given a reference game $(\mathbf{I}, t)$, $S_\text{0}$ embeds $\mathbf{I}_t$ with a convolutional neural network $\fs$, and uses this embedding to initialize a recurrent neural network decoder (RNN-IC) which provides a probability distribution over utterances $u$:
\begin{equation}
\label{eq:s0}
S_\text{0}(u \mid \I, t) = p_{\text{RNN-IC}}(u \mid \fs(I_t)).
\end{equation}
RNN-IC is trained with a standard language modeling objective with teacher forcing: given an image $I$ and ground-truth utterance $u$, the loss of a predicted caption $\hat{u}$ is
\begin{equation}
\mathcal{L}_{S_\text{0}}(\hat{u}, u) = \sum_t p_{\text{RNN-IC}}(\hat{u}_t = u_t \mid u_{<t}; \fs(I_t)).
\end{equation}

\paragraph{Contextual speaker.} The simplest way of incorporating context-sensitivity into $S_\text{0}$ is to condition it on the entire set of images $\I$ as opposed to just $I_t$.
Given the same vision model $f_S$,
let $\C_t$ be the embedding of the target image: $\C_t = \fs(I_t)$.
Let $\C_{t'_1}$ and $\C_{t'_2}$ be embeddings of the distractor images for $t'_i \neq t$: i.e. $\C_{t'_1} = \fs(I_{t'_1})$. Then our \textbf{contextual speaker} $S'_\text{0}$ is trained identically to $S_\text{0}$, except it is conditioned on the concatenation of all three embeddings:
\begin{equation}
S'_\text{0}(u \mid \mathbf{I}, t) = p_{\text{RNN-IC}}(u \mid \mathbf{c}_t; \mathbf{c}_{-t'_1}; \mathbf{c}_{-t'_2} ).
\end{equation}

\subsection{RSA speakers}

RSA speakers generate utterances with the goals of accurate and efficient communication. Accuracy is based on how a pretrained internal listener model $L_0$ will interpret the utterance. 
Given a reference game $\I$ and a speaker utterance $u$, $L_0$ represents the probability of target $t$ as proportional to the dot product between embeddings of $I_t$ and $u$:
\begin{equation}
L_0(t \mid \I,u) \propto \exp \left( \fl(I_t)^{\intercal} g(u) \right),
\end{equation}
where $\fl$ is a CNN encoder (with the same architecture as $\fs$) and $g$ is an RNN encoder.
These encoders will be trained from observed target-utterance pairs.

The speaker $S_\text{RSA}$ then chooses an utterance $u$ which maximizes the probability of the listener identifying $t$, while also balancing the cost of the utterance $C(u)$:
\begin{equation}
\label{eq:rsa}
S_\text{RSA}(u \mid \I, t) \propto \exp \left( \log L_0(t \mid \I, u) - C(u) \right).
\end{equation}
When we can enumerate all possible utterances $\U$, we can compute Eq.~\ref{eq:rsa} exactly, picking $\argmax_{u \in \U} S_\text{RSA}(u \mid \I, t)$; we call this model \textbf{Full RSA} ($S_\text{\text{RSA-Full}}$) and treat it as an upper bound on the pragmatic quality of a speaker model.
However, in many cases, there are an unbounded number of utterances to consider. Past work \cite{andreas2016reasoning,Monroe2017} uses the naive speaker $S_\text{0}$ as a \emph{proposal} distribution from which a subset of $M$ utterances $\U' = (u'_1, \dots, u'_M)$ is sampled; exact inference is then performed with this subset. Since this involves \emph{sampling} candidate utterances from $S_\text{0}$ then \emph{reranking} them based on communicative utility, we refer to these approximations as \textbf{sample-rerank} models ($S_\text{\text{RSA-SRR}}$). For both models, our cost function $C(u)$ is simply a penalty linear in the length of the utterance: $C(u) = \lambda \| u \|$. We used a constant $\lambda = 1$ for $S_\text{\text{RSA-SRR}}$, $\lambda = 0.0001$ for $S_\text{\text{RSA-Full}}$, and $\lambda = 0.01$ for $S_\text{\text{RSA-Am}}$, and for $S_\text{RSA-SRR}$, set $M = 5$.\footnote{Increasing $M$ resulted in little performance benefit.}

\subsection{Learning to produce informative utterances}

Both versions of RSA speakers described so far generate informative utterances by explicitly considering alternatives, which can be slow and expensive.
We next describe a model that \textit{amortizes} this cost, learning to produce the utterances that RSA would prefer, without needing to explicitly consider alternatives at neither train nor test time.

Our \textbf{amortized speaker} $S_\text{RSA-Am}$ has the same architecture as the naive contextual model $S'_\text{0}$, generating an utterance after encoding the target and distractors:
\begin{equation}
\label{eq:am}
S_\text{\text{RSA-Am}}(u \mid \I, t) = p_\text{RNN}(u \mid \mathbf{c}_t; \mathbf{c}_{-t'_1}; \mathbf{c}_{-t'_2} )
\end{equation}
However, while $S'_\text{0}$ is trained as an IC model to match observed utterances, $S_\text{RSA-Am}$ is trained to directly optimize the RSA objective. Specifically, define the loss of a sampled caption $\hat{u}$ as the listener surprisal plus the cost of the utterance:
\begin{equation}
\label{eq:am2}
\mathcal{L}_{S_\text{RSA-Am}}(\hat{u} \mid \I, t) = -\log L_0(t \mid \I, \hat{u}) + C(\hat{u}).
\end{equation}
Equation~\ref{eq:am2} looks similar to the RSA objective, Equation~\ref{eq:rsa}, but crucially omits the normalization term. This means we do not have to consider alternative utterances when training the model:
for a fixed, pre-trained internal listener model $L_0$, each optimization step consists of sampling an utterance from $S_\text{RSA-Am}$, evaluating the listener model, then updating the parameters to minimize $\mathcal{L}_{\text{RSA-Am}}$.
Note that unlike $S'_\text{0}$, utterances are evaluated solely via communicative utility, and not compared to observed language.


A technical problem with training this model is the need to estimate the gradient of $\mathcal{L}_{\text{RSA-Am}}$ given discrete sequences sampled by our speaker. We use the Gumbel-Softmax trick \cite{Jang2017}, which gives differentiable ``samples'' from a categorical distribution by adding Gumbel noise and applying a softmax with a temperature $\tau$. To ensure that $L_0$ receives discrete inputs, we use the (biased) straight-through estimator: the utterance is discretized in the forward pass, but treated as the original continuous sample in the backwards pass. A constant $\tau = 1$ allowed our models to train successfully. Our differentiable cost function $C(u)$ is implemented as a penalty on not predicting the end of sentence token, which increases by $\lambda$ after each sampled token.

$S_\text{RSA-Am}$ has an introspectable internal listener model that can be used for explicit pragmatic reasoning. 
Importantly, this means that the speaker model receives subtle word-by-word supervision during training.
An alternative is an agent that has no internal listener model, but still attempts to maximize communication success by trial and error.
We thus consider a reinforcement learning model, the \textbf{REINFORCE speaker}, $S_{\text{RL}}$, which is architecturally identical to $S_\text{RSA-Am}$, but trained with a black-box reward function rather than an explicit internal listener. A discrete utterance $\hat{u}$ is sampled from $S_{\text{RL}}$ as before.
We simulate a referent choice from an external listener by selecting $\hat{t} = \argmax_t L_0(t \mid \I, \hat{u})$. $S_{\text{RL}}$ receives a reward of +1 if the listener identifies the correct target and 0 otherwise\footnote{This is equivalent to stochastically sampling a reward from the RSA listener with a very low softmax temperature. A higher temperature led to more variance during training and overall worse performance.}. The weights of $S_{\text{RL}}$ are then updated via the REINFORCE gradient estimator \cite{williams1992simple}.
From a reinforcement learning perspective, $S_{\text{RL}}$ and $S_\text{RSA-Am}$ have the same objective, but the former is model-free while the latter is model-based, with RSA as the model of communication.

\section{Experiments}

\subsection{Datasets}

\begin{figure}[t!]
\begin{center}
\includegraphics[width=5cm]{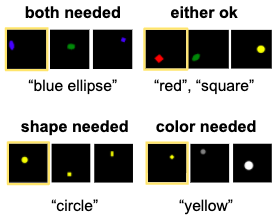}
\end{center}
\caption{Example reference games with ideal utterances.} 
\label{contexts}
\end{figure}

We run experiments on two reference game datasets representing synthetic and human-generated data.
\textbf{ShapeWorld} \cite{Kuhnle2017} is an artificial dataset where each game consists a set of three images, each containing a single random shape with random color. Each game varies in the informativity of the utterance required for successful communication: either only the shape or color is sufficient, or both are needed (Fig.~\ref{contexts}). 
This allows us to evaluate how our speakers modulate their utterances depending on the context. Our primary ShapeWorld dataset has 76000 games containing artificially generated utterances with a total vocabulary of 15 words.
Additionally, to test systematic generalization to novel contexts, we constructed similar datasets where either a color (red), shape (square), or set of color-shape combinations (red circle, blue square, yellow ellipse, white circle, gray square) were held-out during training, but presented as targets in every test game.

\begin{figure*}[t!]
\begin{center}
\includegraphics[width=17.5cm]{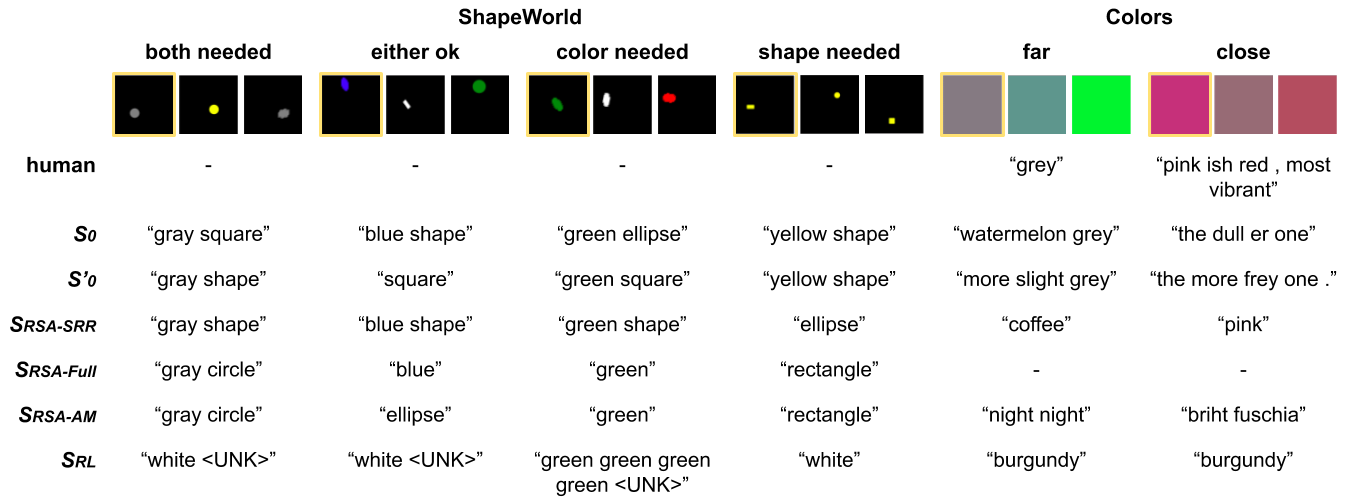}
\end{center}
\caption{Output examples for ShapeWorld and Colors.}
\label{outputs}
\end{figure*}

\textbf{Colors} \cite{Monroe2017} is a reference game dataset with real human language, where human speakers saw three patches of color and were asked to produce utterances that uniquely identify a target color. Here, contexts varied in their difficulty: either distractors were perceptually distinct from the target image (far) or were similar in color space (close) (Fig.~\ref{outputs}).
The language used in the $\sim$46000 games in this dataset is considerably more complex, with around 1400 unique vocabulary tokens; thus, reasoning over all possible utterances (as required by Full RSA) is infeasible.

\subsection{Training and Model Details}
We split each dataset into a training, validation, and test split of 60:15:1 on ShapeWorld and 42:3:1 on Colors. The training datasets for each task were further subdivided into thirds. 1/3 was used to train our speakers (and speaker-internal listener $L_0$ if used). 
The second 1/3 of the training data was used to train a \emph{validation listener}: training was stopped after a speaker obtained maximum accuracy with this listener on the validation set. Finally, the last 1/3 was used to train \emph{evaluation listeners}. The ultimate communicative accuracy of speaker models was evaluated with these evaluation listeners on the test set.
These divisions ensured that speakers did not overfit to a single listener, and instead were evaluated for behavior that generalized across different listeners.


Models were trained for a maximum of 100 epochs with the Adam optimizer \cite{kingma2014adam} with a batch size of 32 and a learning rate of 0.001 for speakers and 0.01 for listeners. After training, the models with the highest validation accuracy were kept. RNNs (both encoders and decoders) are 100-d unidirectional Gated Recurrent Unit (GRU) RNNs \cite{cho2014learning} with 50-d word embeddings learned from scratch, except for the contextual speakers $S'_0$, which have hidden size 300-d (since they consume 3 embeddings). The vision modules $\fs$ and $\fl$ are CNNs with 4 blocks, each block containing a 64-filter 3x3 convolutional layer, batch normalization, ReLU activation, and 2x2 max pooling. Given shapes and colors represented as $64 \times 64 \times 3$ input images, this produces 1024-d representations. In speakers, these are projected via a linear layer down to the GRU hidden size to initialize the GRU; for listeners, to compare image and text embeddings, we project the text representation into the 1024-d image space via a linear layer and use dot product to produce the target probability. Our code is available at \url{https://github.com/juliaiwhite/amortized-rsa}.

\section{Results}

We evaluate our speakers' utterances (see Fig.~\ref{outputs} for examples) in several ways.
First, language should be \textbf{effective}: it should serve its communicative goal of helping the listener correctly identify the target image. We measure communication success with our evaluation listeners on unseen reference games. 
Second, language should be \textbf{concise}, saying as much as needed, but no more. We measure this by examining how the length and content of the utterance changes depending on the difficulty of the contexts as specified in either dataset.
Finally, language should be \textbf{conventional}: to cause minimal confusion to a listener, it should look like conventional, grammatical English. Because the ShapeWorld dataset uses completely artificial language, we measure conventionality on the Colors dataset only.\footnote{RSA and Amortized models used coherent one word utterances, e.g. ``blue'', which due to the artificial nature of the dataset, did not exist in the training data (which had only ``blue shape''). Thus, they had extremely low probability, making this evaluation nonsensical.} As an imperfect proxy to conventionality, we measure the per-word probability of the utterances generated from our speakers, as reported by an unconditional language model trained on utterances in the training data.\footnote{Per-word probability avoids confounds with utterance length.} 
    

\subsubsection{Efficacy.}

\begin{figure}[tbh]
\begin{center}
\includegraphics[width=8.25cm]{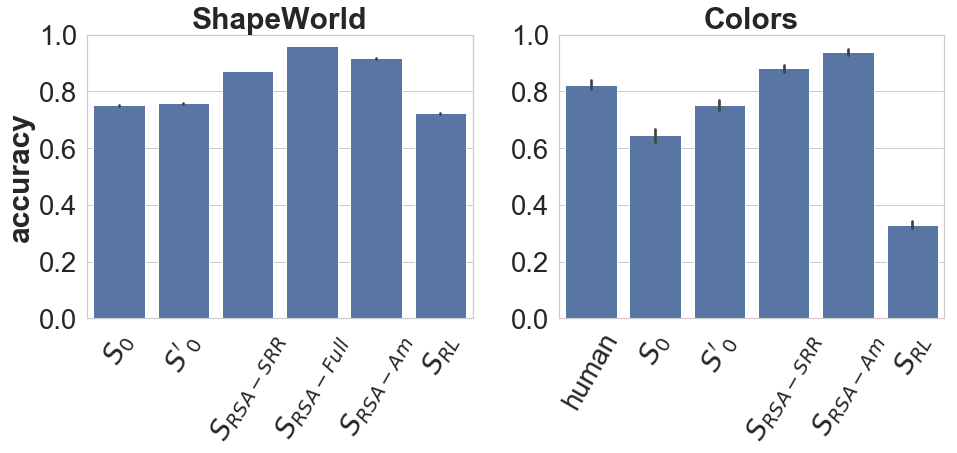}
\end{center}
\caption{\textbf{Efficacy.} Accuracy for ShapeWorld (for 9 evaluation listeners) and Colors (for 1 evaluation listener). Error bars correspond to the 95\% confidence intervals across the test games for all figures.}
\label{efficacy}
\end{figure}

All speaker models, except for the REINFORCE model $S_\text{RL}$, perform well above chance on both test datasets (Fig.~\ref{efficacy}) though there are significant differences between them. 
The naive baseline $S_\text{0}$ attains an accuracy of 73\% for ShapeWorld and 67\% for Colors. While the additional context given to $S'_\text{0}$ results in some improvement, more substantial benefits come when models use pragmatic objectives.
The sample-rerank model $S_\text{RSA-SRR}$ obtains much higher accuracy, only slightly behind $S_\text{RSA-Full}$ in ShapeWorld and even outperforming the human utterances in Colors.\footnote{Human utterance accuracy is imperfect because utterances are evaluated with respect to a neural listener, not a human listener. Our listener model is an imperfect approximation for a human listener, but is still a reasonable measure of efficacy given that we measure accuracy across multiple separately-trained listeners.}
Most notably, our amortized model $S_\text{RSA-Am}$ 
also performs very well, performing on par with $S_\text{RSA-SRR}$ in ShapeWorld and achieving the highest accuracy (94\%)
out of all the models tested for Colors. Meanwhile, the REINFORCE model $S_\text{RL}$ struggles; it performs on par with $S_\text{0}$ for ShapeWorld and is unable to learn in the Colors setting.



\subsubsection{Concision.} 

\begin{figure}[tbh]
\begin{center}
\includegraphics[width=8.25cm]{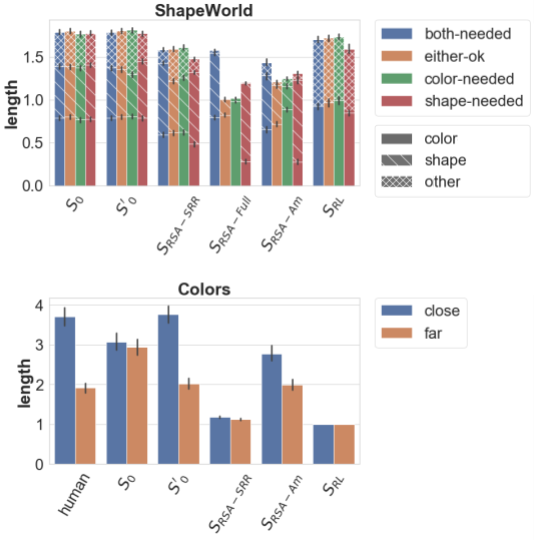}
\end{center}
\caption{\textbf{Concision.} Average utterance lengths produced under different contexts for ShapeWorld and Colors. For Shapeworld, utterance length is split into the average number of colors, shapes, and other tokens (which include the unknown token, "$<$UNK$>$", and generic term "shape") .}
\label{concision}
\end{figure}

We quantify concision by measuring the speaker's average utterance lengths for the different contexts of our reference game (Fig.~\ref{concision}). Analyzing utterance length within context, we see that RSA $S_\text{RSA-Full}$ and the amortized model $S_\text{RSA-Am}$ exhibit appropriately longer utterance lengths when both the shape and color are needed to identify the target image as opposed to either component separately. Looking more specifically at how the number of color and shape words per utterance differ based on the context, we see that in contexts where the shape is required, these models are less likely to say a color and vice versa.\footnote{Both RSA and amortized models also capture the well-attested tendency of humans to produce colors more often than shapes, all else being equal. This in turn follows from properties of the neural listener: a CNN is more accurate at detecting colors than shapes.} This behavior is not reflected at all in the naive speaker $S_\text{0}$, and is less pronounced in the contextual ($S'_\text{0}$) and sample-rerank ($S_\text{RSA-SRR}$) models.
In Colors, we see that humans tailor utterance length to game difficulty: for the difficult ``close'' contexts where colors were the same shade, utterances tended to be longer than the easier ``far'' contexts. Here, only the contextual speaker $S'_\text{0}$ and amortized model $S_\text{RSA-Am}$ demonstrate a similar significant difference in utterance length.

\subsubsection{Conventionality.}

\begin{figure}[tbh]
\begin{center}
\includegraphics[width=8.25cm]{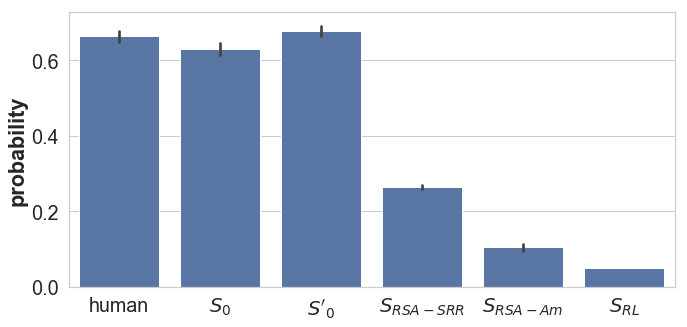}
\end{center}
\caption{\textbf{Conventionality.} Average per-word probability of an utterance for Colors.}
\label{conventionality}
\end{figure}

Finally, we quantify the extent to which generated utterances reflect conventional language, with an unconditional language model trained on human utterances from the Colors dataset. We explore the probability per word of utterances generated by our speaker models (Fig.~\ref{conventionality}).
Unsurprisingly, the naive $S_\text{0}$ and contextual speakers $S'_\text{0}$, which are trained with language-modeling objectives, have high probability. 
The sample re-rank model $S_\text{RSA-SRR}$ sees a major decrease in probability, as it re-ranks sampled utterances with respect to an internal listener, which does not necessarily favor the most probable utterance.

The amortized model $S_\text{RSA-Am}$ is lower still: the communicative training objective results in \emph{language drift} (see Fig.~\ref{outputs} for examples) whereby the model is able to sacrifice realism for communicative efficacy; this has been reported in similarly trained models \cite{Lazaridou2020}. It is unclear whether this reflects creative and acceptable, but unconventional, use of language, or malformed language that would be hard for humans to understand. Regardless, the language is not overfit to the amortized speaker's internal listener, given its ability to generalize to listeners trained on separate data.

\subsubsection{Inference Speed.}

\begin{figure}[tbh]
\begin{center}
\includegraphics[width=8.25cm]{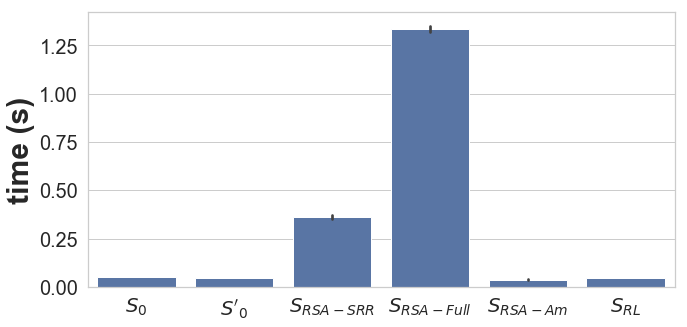}
\end{center}
\caption{\textbf{Inference Speed.} Average time to generate an utterance for ShapeWorld.}
\label{speed}
\end{figure}

\begin{figure}[tbh]
\begin{center}
\includegraphics[width=8.25cm]{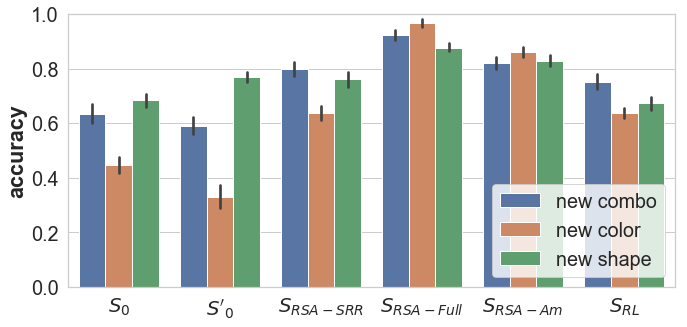}
\end{center}
\caption{\textbf{Generalization.} Accuracy for ShapeWorld when generalizing to a held-out color, shape, or color-shape combination.}
\label{generalization}
\end{figure}

Fig.~\ref{speed} depicts average utterance generation time across 100 games for ShapeWorld. $S_\text{0}$, $S'_\text{0}$, $S_\text{RSA-Am}$, and $S_\text{RL}$ are similarly fast, since they require only a single sample from an RNN. In contrast, $S_\text{RSA-Full}$ and $S_\text{RSA-SRR}$ must evaluate a listener across many utterances, resulting in much higher inference times.
While these results are merely suggestive---our RSA models can likely be optimized---they point to the large gap in compute requirements between online pragmatic reasoning and amortized utterance production.

\subsubsection{Generalization.}

Fig.~\ref{generalization} displays communication accuracy on the dataset variants with held-out colors, shapes, or combinations.
The amortized model comes closest to the performance of Full RSA, followed by the sample-rerank and REINFORCE models.
The contextual captioner's poor performance here likely indicates that, where it captures pragmatic behavior, it does so primarily by memorization. This indicates that communication-based training is needed to produce pragmatic language in novel contexts.


\section{Conclusion and Discussion}

We explored several models for speech production in reference games and evaluated them with respect to pragmatic efficiency, concision, conventionality, and processing time.
The Full RSA model represents the gold standard at a competence level, capturing nuances of pragmatic behavior and generalizing well. 
But it requires large, often intractable, processing costs.
In contrast, our \textit{amortized} RSA model achieves performance close to Full RSA, while being far more efficient.

The contextual speaker, $S'_\text{0}$, performs poorly with respect to efficacy and concision, though well with respect to convention.
Because the contextual model matches our amortized model in architecture but uses a language modeling objective, we conclude that some aspects of pragmatic communication are unattainable by simply observing surface-level linguistic cues.
Despite recent work showing that language models may implicitly learn pragmatic phenomena from sufficient data \cite{schuster2020harnessing}, our simulations suggest that communication-based training is required for successful pragmatic communication, especially when generalizing to situations that have not been seen in the training distribution. The reinforcement learning approach provides a complementary contrast: it shared the communicative objective with the RSA models, but was forced to learn from the weaker signal of communicative success or failure.
The poor performance of this model suggests that an internal model of the listener greatly helps communication-based training.

While common pragmatic behaviors may be amortized, our model must still learn from experience; highly novel situations may still require the full reasoning processes afforded by frameworks like RSA.
Indeed, empirical evidence reports significant variability in processing time across instances of a pragmatic phenomenon \cite{degen2015processing}. To better model this variability, one interesting extension of the work presented here is a model which reverts from amortized computation back to more sophisticated reasoning procedures in novel situations. Such a model would operationalize the trade-off between slow explicit reasoning processes and fast amortized computation that motivates recent theories of language processing \cite{degen2019constraint} and amortized probabilistic inference \cite{gershman2014amortized}.

An intriguing connection is to models of emergent communication \cite{Lazaridou2017,mordatch2018emergence,Lazaridou2020}. 
In order to ground our amortized speaker in actual language, the \emph{listener} is pre-trained on real language and fixed.
In most emergent communication models, either speaker and listener are co-trained from scratch, or in an attempt to ground the communication protocol, the \emph{speaker} is pre-trained and/or given auxiliary language grounding tasks \cite{Lazaridou2017}.

In this paper we described an algorithmic model that forms ``habits of pragmatic speech'', internalizing the explicit pragmatic reasoning of Rational Speech Acts models into a fast but flexible speaker. 
This represents one solution to the puzzle of pragmatic speech: How do we produce pragmatically useful utterances so easily and quickly?


\section{Acknowledgments}

We thank anonymous reviewers and Christopher Potts for insightful comments. This research was supported in part by DARPA under agreement FA8650-19-C-7923, an NSF Graduate Research Fellowship for JM, and the Office of Naval Research grant ONR MURI N00014-16-1-2007.

\bibliographystyle{apacite}

\setlength{\bibleftmargin}{.125in}
\setlength{\bibindent}{-\bibleftmargin}

\bibliography{CogSci_Template}

\begin{thebibliography}{}

\bibitem [\protect \citeauthoryear {%
Andreas%
\ \BBA {} Klein%
}{%
Andreas%
\ \BBA {} Klein%
}{%
{\protect \APACyear {2016}}%
}]{%
andreas2016reasoning}
\APACinsertmetastar {%
andreas2016reasoning}%
\begin{APACrefauthors}%
Andreas, J.%
\BCBT {}\ \BBA {} Klein, D.%
\end{APACrefauthors}%
\unskip\
\newblock
\APACrefYearMonthDay{2016}{}{}.
\newblock
{\BBOQ}\APACrefatitle {Reasoning about Pragmatics with Neural Listeners and
  Speakers} {Reasoning about pragmatics with neural listeners and
  speakers}.{\BBCQ}
\newblock
\BIn{} \APACrefbtitle {{Proceedings of the 2016 Conference on Empirical Methods
  in Natural Language Processing (EMNLP)}} {{Proceedings of the 2016 Conference
  on Empirical Methods in Natural Language Processing (EMNLP)}}\ (\BPGS\
  1173--1182).
\PrintBackRefs{\CurrentBib}

\bibitem [\protect \citeauthoryear {%
Cho%
\ \protect \BOthers {.}}{%
Cho%
\ \protect \BOthers {.}}{%
{\protect \APACyear {2014}}%
}]{%
cho2014learning}
\APACinsertmetastar {%
cho2014learning}%
\begin{APACrefauthors}%
Cho, K.%
, van Merri{\"e}nboer, B.%
, Gulcehre, C.%
, Bahdanau, D.%
, Bougares, F.%
, Schwenk, H.%
\BCBL {}\ \BBA {} Bengio, Y.%
\end{APACrefauthors}%
\unskip\
\newblock
\APACrefYearMonthDay{2014}{}{}.
\newblock
{\BBOQ}\APACrefatitle {Learning Phrase Representations using RNN
  Encoder--Decoder for Statistical Machine Translation} {Learning phrase
  representations using rnn encoder--decoder for statistical machine
  translation}.{\BBCQ}
\newblock
\BIn{} \APACrefbtitle {{Proceedings of the 2014 Conference on Empirical Methods
  in Natural Language Processing (EMNLP)}} {{Proceedings of the 2014 Conference
  on Empirical Methods in Natural Language Processing (EMNLP)}}\ (\BPGS\
  1724--1734).
\PrintBackRefs{\CurrentBib}

\bibitem [\protect \citeauthoryear {%
Cohn-Gordon%
, Goodman%
\BCBL {}\ \BBA {} Potts%
}{%
Cohn-Gordon%
\ \protect \BOthers {.}}{%
{\protect \APACyear {2019}}%
}]{%
Cohn-Gordon2019}
\APACinsertmetastar {%
Cohn-Gordon2019}%
\begin{APACrefauthors}%
Cohn-Gordon, R.%
, Goodman, N.%
\BCBL {}\ \BBA {} Potts, C.%
\end{APACrefauthors}%
\unskip\
\newblock
\APACrefYearMonthDay{2019}{}{}.
\newblock
{\BBOQ}\APACrefatitle {An Incremental Iterated Response Model of Pragmatics}
  {An incremental iterated response model of pragmatics}.{\BBCQ}
\newblock
\BIn{} \APACrefbtitle {{Proceedings of the Society for Computation in
  Linguistics (SCiL) 2019}} {{Proceedings of the Society for Computation in
  Linguistics (SCiL) 2019}}\ (\BPGS\ 81--90).
\PrintBackRefs{\CurrentBib}

\bibitem [\protect \citeauthoryear {%
Degen%
, Hawkins%
, Graf%
, Kreiss%
\BCBL {}\ \BBA {} Goodman%
}{%
Degen%
\ \protect \BOthers {.}}{%
{\protect \APACyear {2020}}%
}]{%
degen2020when}
\APACinsertmetastar {%
degen2020when}%
\begin{APACrefauthors}%
Degen, J.%
, Hawkins, R\BPBI D.%
, Graf, C.%
, Kreiss, E.%
\BCBL {}\ \BBA {} Goodman, N\BPBI D.%
\end{APACrefauthors}%
\unskip\
\newblock
\APACrefYearMonthDay{2020}{}{}.
\newblock
{\BBOQ}\APACrefatitle {When redundancy is useful: A Bayesian approach to
  “overinformative” referring expressions.} {When redundancy is useful: A
  bayesian approach to “overinformative” referring expressions.}{\BBCQ}
\newblock
\APACjournalVolNumPages{Psychological Review}{}{}{}.
\PrintBackRefs{\CurrentBib}

\bibitem [\protect \citeauthoryear {%
Degen%
\ \BBA {} Tanenhaus%
}{%
Degen%
\ \BBA {} Tanenhaus%
}{%
{\protect \APACyear {2015}}%
}]{%
degen2015processing}
\APACinsertmetastar {%
degen2015processing}%
\begin{APACrefauthors}%
Degen, J.%
\BCBT {}\ \BBA {} Tanenhaus, M\BPBI K.%
\end{APACrefauthors}%
\unskip\
\newblock
\APACrefYearMonthDay{2015}{}{}.
\newblock
{\BBOQ}\APACrefatitle {Processing scalar implicature: A constraint-based
  approach} {Processing scalar implicature: A constraint-based
  approach}.{\BBCQ}
\newblock
\APACjournalVolNumPages{Cognitive Science}{39}{4}{667--710}.
\PrintBackRefs{\CurrentBib}

\bibitem [\protect \citeauthoryear {%
Degen%
\ \BBA {} Tanenhaus%
}{%
Degen%
\ \BBA {} Tanenhaus%
}{%
{\protect \APACyear {2019}}%
}]{%
degen2019constraint}
\APACinsertmetastar {%
degen2019constraint}%
\begin{APACrefauthors}%
Degen, J.%
\BCBT {}\ \BBA {} Tanenhaus, M\BPBI K.%
\end{APACrefauthors}%
\unskip\
\newblock
\APACrefYearMonthDay{2019}{}{}.
\newblock
{\BBOQ}\APACrefatitle {Constraint-based pragmatic processing} {Constraint-based
  pragmatic processing}.{\BBCQ}
\newblock
\BIn{} \APACrefbtitle {The {O}xford Handbook of Experimental Semantics and
  Pragmatics.} {The {O}xford handbook of experimental semantics and
  pragmatics.}
\PrintBackRefs{\CurrentBib}

\bibitem [\protect \citeauthoryear {%
Fox%
\ \BBA {} Katzir%
}{%
Fox%
\ \BBA {} Katzir%
}{%
{\protect \APACyear {2011}}%
}]{%
fox2011characterization}
\APACinsertmetastar {%
fox2011characterization}%
\begin{APACrefauthors}%
Fox, D.%
\BCBT {}\ \BBA {} Katzir, R.%
\end{APACrefauthors}%
\unskip\
\newblock
\APACrefYearMonthDay{2011}{}{}.
\newblock
{\BBOQ}\APACrefatitle {On the characterization of alternatives} {On the
  characterization of alternatives}.{\BBCQ}
\newblock
\APACjournalVolNumPages{{Natural Language Semantics}}{19}{1}{87--107}.
\PrintBackRefs{\CurrentBib}

\bibitem [\protect \citeauthoryear {%
Frank%
\ \BBA {} Goodman%
}{%
Frank%
\ \BBA {} Goodman%
}{%
{\protect \APACyear {2012}}%
}]{%
frank2012predicting}
\APACinsertmetastar {%
frank2012predicting}%
\begin{APACrefauthors}%
Frank, M\BPBI C.%
\BCBT {}\ \BBA {} Goodman, N\BPBI D.%
\end{APACrefauthors}%
\unskip\
\newblock
\APACrefYearMonthDay{2012}{}{}.
\newblock
{\BBOQ}\APACrefatitle {Predicting pragmatic reasoning in language games}
  {Predicting pragmatic reasoning in language games}.{\BBCQ}
\newblock
\APACjournalVolNumPages{Science}{336}{6084}{998--998}.
\PrintBackRefs{\CurrentBib}

\bibitem [\protect \citeauthoryear {%
Gershman%
\ \BBA {} Goodman%
}{%
Gershman%
\ \BBA {} Goodman%
}{%
{\protect \APACyear {2014}}%
}]{%
gershman2014amortized}
\APACinsertmetastar {%
gershman2014amortized}%
\begin{APACrefauthors}%
Gershman, S.%
\BCBT {}\ \BBA {} Goodman, N.%
\end{APACrefauthors}%
\unskip\
\newblock
\APACrefYearMonthDay{2014}{}{}.
\newblock
{\BBOQ}\APACrefatitle {Amortized inference in probabilistic reasoning}
  {Amortized inference in probabilistic reasoning}.{\BBCQ}
\newblock
\BIn{} \APACrefbtitle {{Proceedings of the Annual Meeting of the Cognitive
  Science Society (CogSci)}} {{Proceedings of the Annual Meeting of the
  Cognitive Science Society (CogSci)}}\ (\BVOL~36).
\PrintBackRefs{\CurrentBib}

\bibitem [\protect \citeauthoryear {%
Goodman%
\ \BBA {} Frank%
}{%
Goodman%
\ \BBA {} Frank%
}{%
{\protect \APACyear {2016}}%
}]{%
Goodman2016}
\APACinsertmetastar {%
Goodman2016}%
\begin{APACrefauthors}%
Goodman, N\BPBI D.%
\BCBT {}\ \BBA {} Frank, M\BPBI C.%
\end{APACrefauthors}%
\unskip\
\newblock
\APACrefYearMonthDay{2016}{}{}.
\newblock
{\BBOQ}\APACrefatitle {Pragmatic language interpretation as probabilistic
  inference} {Pragmatic language interpretation as probabilistic
  inference}.{\BBCQ}
\newblock
\APACjournalVolNumPages{Trends in Cognitive Sciences}{20}{11}{818-829}.
\PrintBackRefs{\CurrentBib}

\bibitem [\protect \citeauthoryear {%
Grice%
}{%
Grice%
}{%
{\protect \APACyear {1975}}%
}]{%
Grice1975}
\APACinsertmetastar {%
Grice1975}%
\begin{APACrefauthors}%
Grice, H\BPBI P.%
\end{APACrefauthors}%
\unskip\
\newblock
\APACrefYearMonthDay{1975}{}{}.
\newblock
{\BBOQ}\APACrefatitle {Logic and Conversation} {Logic and conversation}.{\BBCQ}
\newblock
\BIn{} P.~Cole\ \BBA {} J\BPBI L.~Morgan\ (\BEDS), \APACrefbtitle {Syntax and
  Semantics: Vol. 3: Speech Acts.} {Syntax and semantics: Vol. 3: Speech acts.}
\newblock
\APACaddressPublisher{New York}{Academic Press}.
\PrintBackRefs{\CurrentBib}

\bibitem [\protect \citeauthoryear {%
Jang%
, Gu%
\BCBL {}\ \BBA {} Poole%
}{%
Jang%
\ \protect \BOthers {.}}{%
{\protect \APACyear {2017}}%
}]{%
Jang2017}
\APACinsertmetastar {%
Jang2017}%
\begin{APACrefauthors}%
Jang, E.%
, Gu, S.%
\BCBL {}\ \BBA {} Poole, B.%
\end{APACrefauthors}%
\unskip\
\newblock
\APACrefYearMonthDay{2017}{}{}.
\newblock
{\BBOQ}\APACrefatitle {Categorical Reparameterization with Gumbel-Softmax}
  {Categorical reparameterization with gumbel-softmax}.{\BBCQ}
\newblock
\BIn{} \APACrefbtitle {{International Conference on Learning Representations
  (ICLR)}.} {{International Conference on Learning Representations (ICLR)}.}
\PrintBackRefs{\CurrentBib}

\bibitem [\protect \citeauthoryear {%
Kingma%
\ \BBA {} Ba%
}{%
Kingma%
\ \BBA {} Ba%
}{%
{\protect \APACyear {2014}}%
}]{%
kingma2014adam}
\APACinsertmetastar {%
kingma2014adam}%
\begin{APACrefauthors}%
Kingma, D\BPBI P.%
\BCBT {}\ \BBA {} Ba, J.%
\end{APACrefauthors}%
\unskip\
\newblock
\APACrefYearMonthDay{2014}{}{}.
\newblock
{\BBOQ}\APACrefatitle {Adam: A method for stochastic optimization} {Adam: A
  method for stochastic optimization}.{\BBCQ}
\newblock
\BIn{} \APACrefbtitle {{International Conference on Learning Representations
  (ICLR)}.} {{International Conference on Learning Representations (ICLR)}.}
\PrintBackRefs{\CurrentBib}

\bibitem [\protect \citeauthoryear {%
Krahmer%
\ \BBA {} Van~Deemter%
}{%
Krahmer%
\ \BBA {} Van~Deemter%
}{%
{\protect \APACyear {2012}}%
}]{%
krahmer2012computational}
\APACinsertmetastar {%
krahmer2012computational}%
\begin{APACrefauthors}%
Krahmer, E.%
\BCBT {}\ \BBA {} Van~Deemter, K.%
\end{APACrefauthors}%
\unskip\
\newblock
\APACrefYearMonthDay{2012}{}{}.
\newblock
{\BBOQ}\APACrefatitle {Computational generation of referring expressions: A
  survey} {Computational generation of referring expressions: A survey}.{\BBCQ}
\newblock
\APACjournalVolNumPages{Computational Linguistics}{38}{1}{173--218}.
\PrintBackRefs{\CurrentBib}

\bibitem [\protect \citeauthoryear {%
Kuhnle%
\ \BBA {} Copestake%
}{%
Kuhnle%
\ \BBA {} Copestake%
}{%
{\protect \APACyear {2017}}%
}]{%
Kuhnle2017}
\APACinsertmetastar {%
Kuhnle2017}%
\begin{APACrefauthors}%
Kuhnle, A.%
\BCBT {}\ \BBA {} Copestake, A.%
\end{APACrefauthors}%
\unskip\
\newblock
\APACrefYearMonthDay{2017}{}{}.
\newblock
{\BBOQ}\APACrefatitle {Shape{W}orld - A new test methodology for multimodal
  language understanding} {Shape{W}orld - a new test methodology for multimodal
  language understanding}.{\BBCQ}
\newblock
\APACjournalVolNumPages{arXiv preprint}{}{}{}.
\newblock
\begin{APACrefDOI} \doi{arXiv:1704.04517} \end{APACrefDOI}
\PrintBackRefs{\CurrentBib}

\bibitem [\protect \citeauthoryear {%
Lakoff%
\ \BBA {} Johnson%
}{%
Lakoff%
\ \BBA {} Johnson%
}{%
{\protect \APACyear {2008}}%
}]{%
lakoff2008metaphors}
\APACinsertmetastar {%
lakoff2008metaphors}%
\begin{APACrefauthors}%
Lakoff, G.%
\BCBT {}\ \BBA {} Johnson, M.%
\end{APACrefauthors}%
\unskip\
\newblock
\APACrefYear{2008}.
\newblock
\APACrefbtitle {Metaphors we live by} {Metaphors we live by}.
\newblock
\APACaddressPublisher{}{University of Chicago Press}.
\PrintBackRefs{\CurrentBib}

\bibitem [\protect \citeauthoryear {%
Lazaridou%
, Peysakhovich%
\BCBL {}\ \BBA {} Baroni%
}{%
Lazaridou%
\ \protect \BOthers {.}}{%
{\protect \APACyear {2017}}%
}]{%
Lazaridou2017}
\APACinsertmetastar {%
Lazaridou2017}%
\begin{APACrefauthors}%
Lazaridou, A.%
, Peysakhovich, A.%
\BCBL {}\ \BBA {} Baroni, M.%
\end{APACrefauthors}%
\unskip\
\newblock
\APACrefYearMonthDay{2017}{}{}.
\newblock
{\BBOQ}\APACrefatitle {Multi-Agent Cooperation and the Emergence of (Natural)
  Language} {Multi-agent cooperation and the emergence of (natural)
  language}.{\BBCQ}
\newblock
\BIn{} \APACrefbtitle {{International Conference on Learning Representations
  (ICLR)}.} {{International Conference on Learning Representations (ICLR)}.}
\PrintBackRefs{\CurrentBib}

\bibitem [\protect \citeauthoryear {%
Lazaridou%
, Potapenko%
\BCBL {}\ \BBA {} Tieleman%
}{%
Lazaridou%
\ \protect \BOthers {.}}{%
{\protect \APACyear {2020}}%
}]{%
Lazaridou2020}
\APACinsertmetastar {%
Lazaridou2020}%
\begin{APACrefauthors}%
Lazaridou, A.%
, Potapenko, A.%
\BCBL {}\ \BBA {} Tieleman, O.%
\end{APACrefauthors}%
\unskip\
\newblock
\APACrefYearMonthDay{2020}{}{}.
\newblock
{\BBOQ}\APACrefatitle {Multi-agent Communication meets Natural Language:
  Synergies between Functional and Structural Language Learning} {Multi-agent
  communication meets natural language: Synergies between functional and
  structural language learning}.{\BBCQ}
\newblock
\BIn{} \APACrefbtitle {{Proceedings of the 58th Annual Meeting of the
  Association for Computational Linguistics (ACL)}.} {{Proceedings of the 58th
  Annual Meeting of the Association for Computational Linguistics (ACL)}.}
\PrintBackRefs{\CurrentBib}

\bibitem [\protect \citeauthoryear {%
Levinson%
}{%
Levinson%
}{%
{\protect \APACyear {2000}}%
}]{%
levinson2000presumptive}
\APACinsertmetastar {%
levinson2000presumptive}%
\begin{APACrefauthors}%
Levinson, S\BPBI C.%
\end{APACrefauthors}%
\unskip\
\newblock
\APACrefYear{2000}.
\newblock
\APACrefbtitle {Presumptive meanings: The theory of generalized conversational
  implicature} {Presumptive meanings: The theory of generalized conversational
  implicature}.
\newblock
\APACaddressPublisher{}{MIT Press}.
\PrintBackRefs{\CurrentBib}

\bibitem [\protect \citeauthoryear {%
Lewis%
}{%
Lewis%
}{%
{\protect \APACyear {1969}}%
}]{%
Lewis1969}
\APACinsertmetastar {%
Lewis1969}%
\begin{APACrefauthors}%
Lewis, D.%
\end{APACrefauthors}%
\unskip\
\newblock
\APACrefYear{1969}.
\newblock
\APACrefbtitle {Convention: A Philosophical Study} {Convention: A philosophical
  study}.
\newblock
\APACaddressPublisher{Cambridge, MA}{Harvard University Press}.
\PrintBackRefs{\CurrentBib}

\bibitem [\protect \citeauthoryear {%
Marr%
\ \BBA {} Poggio%
}{%
Marr%
\ \BBA {} Poggio%
}{%
{\protect \APACyear {1976}}%
}]{%
marr1976understanding}
\APACinsertmetastar {%
marr1976understanding}%
\begin{APACrefauthors}%
Marr, D.%
\BCBT {}\ \BBA {} Poggio, T.%
\end{APACrefauthors}%
\unskip\
\newblock
\APACrefYear{1976}.
\newblock
\APACrefbtitle {From understanding computation to understanding neural
  circuitry} {From understanding computation to understanding neural
  circuitry}.
\newblock
\APACaddressPublisher{}{Massachusetts Institute of Technology, Artificial
  Intelligence Laboratory}.
\PrintBackRefs{\CurrentBib}

\bibitem [\protect \citeauthoryear {%
Monroe%
, Hawkins%
, Goodman%
\BCBL {}\ \BBA {} Potts%
}{%
Monroe%
\ \protect \BOthers {.}}{%
{\protect \APACyear {2017}}%
}]{%
Monroe2017}
\APACinsertmetastar {%
Monroe2017}%
\begin{APACrefauthors}%
Monroe, W.%
, Hawkins, R\BPBI X\BPBI D.%
, Goodman, N\BPBI D.%
\BCBL {}\ \BBA {} Potts, C.%
\end{APACrefauthors}%
\unskip\
\newblock
\APACrefYearMonthDay{2017}{}{}.
\newblock
{\BBOQ}\APACrefatitle {Colors in Context: A Pragmatic Neural Model for Grounded
  Language Understanding} {Colors in context: A pragmatic neural model for
  grounded language understanding}.{\BBCQ}
\newblock
\APACjournalVolNumPages{Transactions of the Association for Computational
  Linguistics (TACL)}{5}{}{325--338}.
\PrintBackRefs{\CurrentBib}

\bibitem [\protect \citeauthoryear {%
Mordatch%
\ \BBA {} Abbeel%
}{%
Mordatch%
\ \BBA {} Abbeel%
}{%
{\protect \APACyear {2018}}%
}]{%
mordatch2018emergence}
\APACinsertmetastar {%
mordatch2018emergence}%
\begin{APACrefauthors}%
Mordatch, I.%
\BCBT {}\ \BBA {} Abbeel, P.%
\end{APACrefauthors}%
\unskip\
\newblock
\APACrefYearMonthDay{2018}{}{}.
\newblock
{\BBOQ}\APACrefatitle {Emergence of grounded compositional language in
  multi-agent populations} {Emergence of grounded compositional language in
  multi-agent populations}.{\BBCQ}
\newblock
\BIn{} \APACrefbtitle {{Thirty-Second AAAI Conference on Artificial
  Intelligence}.} {{Thirty-Second AAAI Conference on Artificial Intelligence}.}
\PrintBackRefs{\CurrentBib}

\bibitem [\protect \citeauthoryear {%
Schuster%
, Chen%
\BCBL {}\ \BBA {} Degen%
}{%
Schuster%
\ \protect \BOthers {.}}{%
{\protect \APACyear {2020}}%
}]{%
schuster2020harnessing}
\APACinsertmetastar {%
schuster2020harnessing}%
\begin{APACrefauthors}%
Schuster, S.%
, Chen, Y.%
\BCBL {}\ \BBA {} Degen, J.%
\end{APACrefauthors}%
\unskip\
\newblock
\APACrefYearMonthDay{2020}{}{}.
\newblock
{\BBOQ}\APACrefatitle {Harnessing the richness of the linguistic signal in
  predicting pragmatic inferences} {Harnessing the richness of the linguistic
  signal in predicting pragmatic inferences}.{\BBCQ}
\newblock
\BIn{} \APACrefbtitle {{Proceedings of the 58th Annual Meeting of the
  Association for Computational Linguistics (ACL)}.} {{Proceedings of the 58th
  Annual Meeting of the Association for Computational Linguistics (ACL)}.}
\PrintBackRefs{\CurrentBib}

\bibitem [\protect \citeauthoryear {%
Searle%
}{%
Searle%
}{%
{\protect \APACyear {1969}}%
}]{%
searle1969speech}
\APACinsertmetastar {%
searle1969speech}%
\begin{APACrefauthors}%
Searle, J\BPBI R.%
\end{APACrefauthors}%
\unskip\
\newblock
\APACrefYear{1969}.
\newblock
\APACrefbtitle {Speech acts: An essay in the philosophy of language} {Speech
  acts: An essay in the philosophy of language}\ (\BVOL~626).
\newblock
\APACaddressPublisher{}{Cambridge University Press}.
\PrintBackRefs{\CurrentBib}

\bibitem [\protect \citeauthoryear {%
Sperber%
\ \BBA {} Wilson%
}{%
Sperber%
\ \BBA {} Wilson%
}{%
{\protect \APACyear {1986}}%
}]{%
sperber1986relevance}
\APACinsertmetastar {%
sperber1986relevance}%
\begin{APACrefauthors}%
Sperber, D.%
\BCBT {}\ \BBA {} Wilson, D.%
\end{APACrefauthors}%
\unskip\
\newblock
\APACrefYear{1986}.
\newblock
\APACrefbtitle {Relevance: Communication and cognition} {Relevance:
  Communication and cognition}\ (\BVOL~142).
\newblock
\APACaddressPublisher{Cambridge, MA}{Harvard University Press}.
\PrintBackRefs{\CurrentBib}

\bibitem [\protect \citeauthoryear {%
van Gompel%
, van Deemter%
, Gatt%
, Snoeren%
\BCBL {}\ \BBA {} Krahmer%
}{%
van Gompel%
\ \protect \BOthers {.}}{%
{\protect \APACyear {2019}}%
}]{%
van2019conceptualization}
\APACinsertmetastar {%
van2019conceptualization}%
\begin{APACrefauthors}%
van Gompel, R\BPBI P.%
, van Deemter, K.%
, Gatt, A.%
, Snoeren, R.%
\BCBL {}\ \BBA {} Krahmer, E\BPBI J.%
\end{APACrefauthors}%
\unskip\
\newblock
\APACrefYearMonthDay{2019}{}{}.
\newblock
{\BBOQ}\APACrefatitle {Conceptualization in reference production: Probabilistic
  modeling and experimental testing.} {Conceptualization in reference
  production: Probabilistic modeling and experimental testing.}{\BBCQ}
\newblock
\APACjournalVolNumPages{Psychological Review}{126}{3}{345}.
\PrintBackRefs{\CurrentBib}

\bibitem [\protect \citeauthoryear {%
Williams%
}{%
Williams%
}{%
{\protect \APACyear {1992}}%
}]{%
williams1992simple}
\APACinsertmetastar {%
williams1992simple}%
\begin{APACrefauthors}%
Williams, R\BPBI J.%
\end{APACrefauthors}%
\unskip\
\newblock
\APACrefYearMonthDay{1992}{}{}.
\newblock
{\BBOQ}\APACrefatitle {Simple statistical gradient-following algorithms for
  connectionist reinforcement learning} {Simple statistical gradient-following
  algorithms for connectionist reinforcement learning}.{\BBCQ}
\newblock
\APACjournalVolNumPages{Machine learning}{8}{3-4}{229--256}.
\PrintBackRefs{\CurrentBib}

\end{thebibliography}

\end{document}